# Predictive Analysis of Diabetic Retinopathy with Transfer Learning

Raj Sunil Salvi
*Department of Computer Engineering*
Fr. Conceicao Rodrigues Institute of Technology, Vashi
Navi Mumbai, India
raj.salvi@fcrit.onmicrosoft.com

Shreyas Rajesh Labhsetwar
*Department of Computer Engineering*
Fr. Conceicao Rodrigues Institute of Technology, Vashi
Navi Mumbai, India
shreyas.labhsetwar@fcrit.onmicrosoft.com

Piyush Arvind Kolte
*Department of Computer Engineering*
Fr. Conceicao Rodrigues Institute of Technology, Vashi
Navi Mumbai, India
piyush.kolte@fcrit.onmicrosoft.com

Veerasai Subramaniam venkatesh
*Department of Computer Engineering*
Fr. Conceicao Rodrigues Institute of Technology, Vashi
Navi Mumbai, India
veerasai.subramaniam@fcrit.onmicrosoft.com

Alistair Michael Baretto
*Department of Computer Engineering*
Fr. Conceicao Rodrigues Institute of Technology, Vashi
Navi Mumbai, India
alistair.baretto@fcrit.onmicrosoft.com

*Abstract*— With the prevalence of diabetes, Diabetes Mellitus Retinopathy (DR) is becoming a major health problem across the world. The long-term medical complications arising due to DR significantly affect the patient as well as the society, as the disease mostly affects young and productive individuals. Early detection and treatment can help reduce the extent of damage to the patients. The rise of Convolutional Neural Networks for predictive analysis in the medical field paves the way for a robust solution to DR detection. This paper studies the performance of several highly efficient and scalable CNN architectures for Diabetic Retinopathy Classification with the help of Transfer Learning. The research focuses on VGG16, Resnet50 V2, and EfficientNet B0 models. The classification performance is analyzed using several performance measures including True Positive Rate, False Positive Rate, Accuracy, etc. Also, several performance graphs are plotted for visualizing the architecture performance including Confusion Matrix, ROC Curve, etc. The results indicate that Transfer Learning with ImageNet weights using VGG 16 model demonstrates the best classification performance with an accuracy of 95%. It is closely followed by ResNet50 V2 architecture with an accuracy of 93%. This paper shows that predictive analysis of DR from retinal images is achieved with Transfer Learning on Convolutional Neural Networks.

## I. Introduction

Diabetic retinopathy (DR) comprises an eye sickness that often affects individuals with diabetes, and may lead to total blindness if not treated timely. The back of the human eye is covered with a membrane called Retina, which is highly sensitive to light. The conversion of light into electrical signals is performed by the retina, and these signals are carried to the brain by the Nervous System. This process produces visual images, enabling the sense of sight in the human eye. Diabetic retinopathy results in leakage of fluid from the damaged blood vessels within the retinal tissue, causing distortion in vision. The analysis of the degree and severity of retinopathy in a patient is presently performed by doctors with the help of fundus retinal scans. However, this task is laborious, takes time, and demands a lot of effort because of the minuscule dimensions of the lesions and their obscurity in contrast. To overcome this difficulty, a deep-learning-based solution can be implemented.

Through the advancements of neural networks in the form of convolutional neural networks, classification problems can be easily solved by detecting the core patterns in the images, thereby mapping them to their respective classes. Due to the existence of larger data sets, sophisticated models and training algorithms, and high computational power with GPUs, larger and deeper models can be easily trained. Convolutional Networks have shown excellence in performing tasks such as hand-written digit classification and face detection. Several papers have demonstrated their potential for outstanding performance on more challenging visual classification tasks.

In order to perform the task of DR classification, we have used the transfer learning methodology to train classifiers on fundus retinal images on the following well-known architectures, namely, VGG16, ResNet50 V2, and EffcientNet B0, followed by a detailed comparison of their graded classification performance and selecting the best architecture. Based on the severity of the damage caused by Diabetic Retinopathy, classification is performed into the following classes, viz. mild, moderate, No DR, Proliferate DR, and Severe DR.

## II. Related work

Lifeng Qiao et al. [1] proposed a deep learning methodology that predicts the occurrence of microaneurysm from fundus scans for early identification of diabetic retinopathy. They have implemented semantic segmentation to aid ophthalmologists to classify as early, moderate and severe NDPR.

Farrikh Alzami et al. [2] perform diabetic retinopathy grade classification using fractal dimension features, which helps them characterize the retinal vasculature. They performing Supervised Learning with Random Forest Classifier on the MESSIDOR dataset, and conclude that fractal dimensions help classify the healthy and diabetic retinopathy images. Still, the graded classification remains a challenge.

Mohamed Chetoui ae al. [3] implemented a CNN model using transfer learning on Google's EfficientNet for detecting referable and vision-threatening DR. They used two datasets, EyePACS and APTOS for their research, and achieved a high area under curve scores of 0.984 for referable diabetic retinopathy images, whilst 0.990 for vision-threatening DR images. Further, they developed an explainability algorithm demonstrating the efficacy of their algorithm in DR detection.

Asra Momeni Pour et al. [4] propose a deep learning methodology involving Contrast Limited Adaptive Histogram Equalization algorithm for enhancing the image quality comprising the dataset and uniform equalization of their intensities for preprocessing. The dataset is used to train the EfficientNet-B5 CNN model which achieves a higher classification accuracy by scaling the model in width, depth, and resolution dimensions. The research demonstrated an excellent classification performance with an AUC score of 0.945.

Ling Shao et al. [5] and Huan Liang et al. [6] analyze the feature space similarities in training and future datasets, and how it impacts the prediction capability of the Neural Networks. They discover that transfer learning resolves cross-domain learning problems by extracting all the insights from data in a related domain and transfer them for use in target tasks. They perform a detailed survey of various state-of-the-art transfer learning algorithms in visual categorization algorithms, such as object recognition, image classification, and human action recognition.

Mohsen Hajabdollahi et al. [7] explore reducing the structural complexity of CNN architectures for DR analysis by hierarchical pruning methodology. They implement a modified VGG16 model with fewer parameters by eliminating useless connections, filter channels, and filters to reduce the complexity of the network structure. This research concluded that hierarchical pruning on the VGG16 model resulted in the elimination of 35% feature maps with just a 1.89% accuracy drop.

## III. PROPOSED WORK

This paper designs, implements, compares, and selects the best model for performing the task of severity detection (grading) of retinopathy in a diabetic patient. This task is achieved by training the dataset on the following CNN models: VGG16 model, ResNet50 V2 model, and EfficientNet B0 model.

The presented research analyses the performance of different ConvNets on the specified dataset and also prescribes the perfect pre-trained network for the task of diabetic retinopathy classification. To do so, CNN models are trained on the DR Images dataset with transfer learning, which consists of leveraging the stored information (knowledge) learned while solving one task and applying it to a different, semantically related task.

### A. Dataset Description

The dataset [8] that we are using is an open-source diabetic retinopathy detection dataset comprising of 1000 images, split in a ratio of 80:20, meaning, 800 images comprise the training dataset and 200 images comprise the validation dataset. The classification is performed into 5 classes, viz. Mild DR, Moderate DR, No DR, Proliferate DR, and Severe DR. A sample of the images from the corresponding classes is shown in Fig. 1.

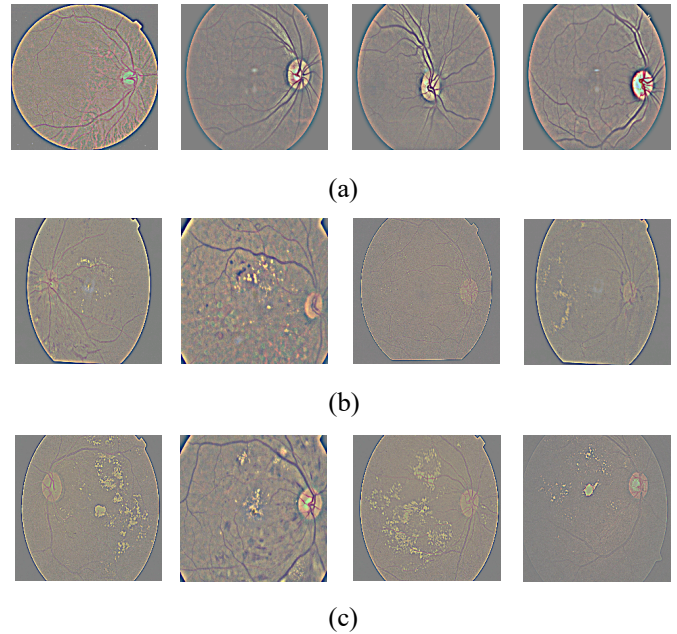

(a)

(b)

(c)

Fig. 1. (a) No DR, (b) Proliferate DR, (c) Severe DR Images

### B. Data Preprocessing

Since deep learning requires a bulk of labelled images for the classification tasks, the size of the dataset is not enough to perform transfer learning on state-of-the-art CNN architectures. Hence, several data augmentation techniques are applied, including cropping, shearing, rotation, and flipping [9]. After this task, the data is normalized, since a normalized distribution of the dataset ensures faster convergence while training. Since the dataset is small, the training task is performed with the data in a single batch.

### C. Network Architecture

For the purpose of image classification, we are using convolutional neural networks that employ deep learning methodology for classification problems [10]. They have a layered structure with the following layer types:

1. Convolution Layer
2. Sub-sampling Layer (Max Pooling Layer)
3. Full Connection Layer

The research involves training the dataset with the help of transfer learning using major pre-trained networks like VGG16, ResNet50 V2, and EfficientNet B0. These networks are designed to train some of the well-known and huge datasets viz. ImageNet, Cifar-10, etc. In order to have a common base for comparison, we use the aforementioned models with pre-trained weights for ImageNet (Dataset of over fifteen million high-resolution labeled images belonging to over 22,000 categories) [11].

## D. VGG16 Model

VGG16 is short for Visual Geometry Group. It is a convolutional neural network model, implemented by K. Simonyan and A. Zisserman [12].

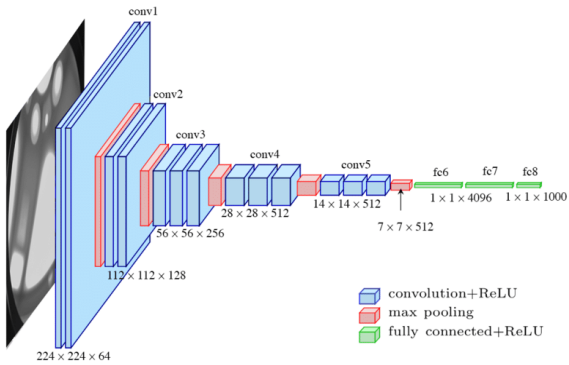

Fig. 2. VGG16 Model

The VGG16 model's architecture, as displayed in Fig. 2, consists of total 12 layers. The first Conv layer is of fixed size with an input image resolution of (224 X 224 X 3). This image is propagated through a set of convolutional layers. Due to its larger width and depth, the pre-trained weights of a VGG16 model is about 500 MB in size.

## E. ResNet50 V2 Model

ResNet, which is short for Residual Networks, is a classic neural network used in many image classification and computer vision tasks. Resnet has a depth of 152 layers, as shown in Fig. 3. We are using a ResNet50 model which is 50 layers deep and has a special connection-skipping property, which helps overcome the problem of vanishing gradients in bigger models [13].

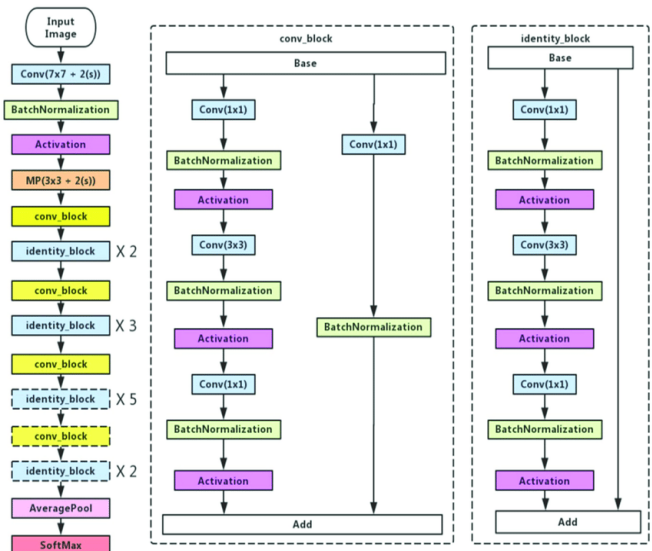

Fig. 3. ResNet50 V2 Architecture

## F. EfficientNet Model

EfficientNet is one of the most efficient CNN models, whilst needing the least FLOPS for inference, and exhibits brilliant accuracy on both ImageNet and regular image classification with transfer learning tasks. This paper utilises a baseline EfficientNet Model, i.e., EfficientNet B0 model with input image size of (224 X 224 X 3) [14].

| Stage | Operator | Resolution | #channels | #layers |
|---|---|---|---|---|
| 1 | Conv3x3 | 224x224 | 32 | 1 |
| 2 | MBConv1,k3x3 | 112x112 | 16 | 1 |
| 3 | MBConv6,k3x3 | 112x112 | 24 | 2 |
| 4 | MBConv6,k5x5 | 56x56 | 40 | 2 |
| 5 | MBConv6,k3x3 | 28x28 | 80 | 3 |
| 6 | MBConv6,k5x5 | 14x14 | 112 | 3 |
| 7 | MBConv6,k5x5 | 14x14 | 192 | 4 |
| 8 | MBConv6,k3x3 | 7x7 | 320 | 1 |
| 9 | Conv1x1/Pooling/FC | 7x7 | 1, 280 | 1 |

Fig. 4. EfficientNet Architecture

## G. Transfer Learning

Transfer learning is a machine learning algorithm involving utilizing a model developed for a particular task as the base for a model on a different but related second task [15]. It is preferred in deep learning tasks involving using existing Deep Learning models as the base for CV and NLP tasks to cut down on the vast computational and temporal resources necessary to build neural network algorithms on these problems.

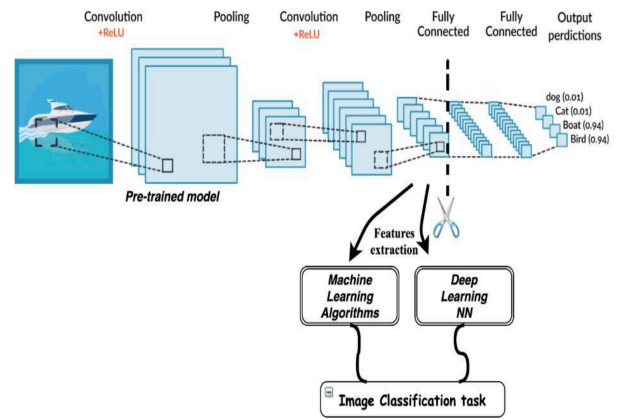

Fig. 5. Transfer Learning

This paper uses the models available for download via Keras.application package of TensorFlow. Since we use the learned knowledge stored in the form of weights for the above-mentioned models trained on the ImageNet dataset, there is no need for training the deeper layers, thus saving time and energy. To do so, TensorFlow allows to exclude the layer top which is the classification layer of the respective models, and also allows freezing the weights of the remaining layers, thus avoiding retraining. We append a stack of 4 dense layers with depths of 256,128,128, and 5 nodes respectively with the last layer being the classification layer with the activation function for the first four layers as ReLu and SoftMax function for the last layer [16].

### H. Training Process

The training process involves converting all the CNN layers to a sequential linear stack by using the sequential model API available in Keras. The individual networks are trained for 15 epochs with a call-back set for monitoring the validation accuracy, thus adjusting the rate of learning with a factor of 0.5 units, each time the validation accuracy begins getting stagnant [17].

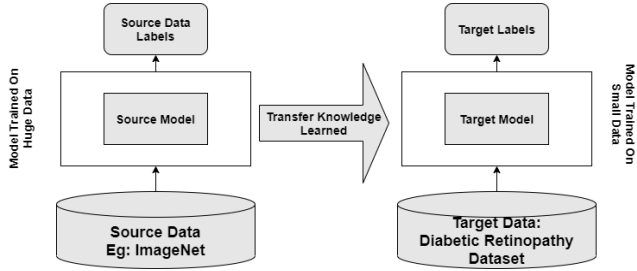

Fig. 6. Training Process

### I. Adam Optimizer

Optimizers are algorithms that help compute errors upon forward-propagations and thus help in adjusting the parameters of a neural network including its weights and learning rate to lower the losses. This research involves using Adam optimizer for all the networks to have a common base for training purposes. Adam Optimizer introduces the property of momentum in the RMSprop Optimizer. It regulates the gradient component with respect to the exponential moving average of gradients ($\mu$) and the learning rate component by dividing the learning rate $\alpha$ by $\sqrt{\lambda}$, the exponential moving average of squared gradients (similar to RMSprop).

### J. Loss Function

Training neural networks involve using an optimization process that utilizes a loss function for the calculation of the model error. A loss function, in simple terms, is an object function, whose minima is to be determined. Depending upon the learning task, Loss functions are classified into two major categories, namely, Regression Loss functions and Classification Loss functions [20]. This research involves making use of a classification loss function called categorical cross-entropy loss due to the multiclass nature of the DR classification problem.

## IV. RESULTS

This research involves transfer learning on all the aforementioned CNN models with the open-source Diabetic Retinopathy Classification Dataset. The performance is measured with various performance measures including Precision, Sensitivity, F1-Score, etc. Also, performance graphs including Training and Validation Accuracy and Loss curves, ROC curves, and Confusion Matrix are plotted for each CNN architecture trained for 15 epochs using the Adam Optimizer.

### A. EfficientNet B0 Architecture

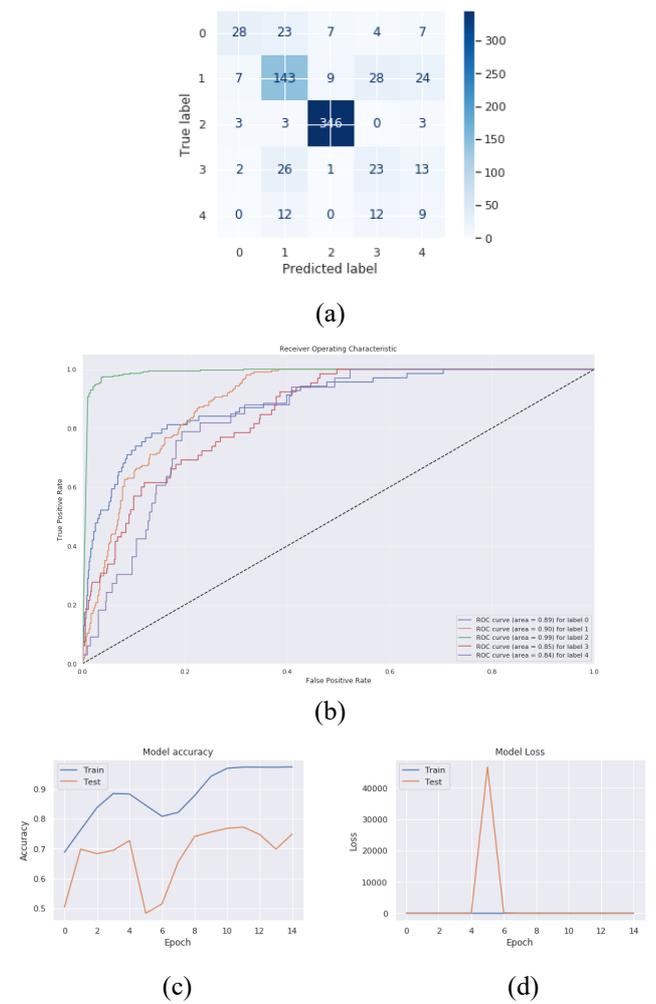

Fig. 7. (a) Confusion Matrix (b) ROC curve (c) Training and Validation Accuracy Curve (d) Training and Validation Loss Curve

From Fig. 7(a), we can observe that EfficientNet B0 misclassifies several Mild DR images as Moderate DR images, Proliferate DR images as Moderate DR images, and Severe DR images as Moderate and Proliferate DR images. Fig. 7(b) demonstrates that the best classification performance is achieved for Non-DR images. From Fig. 7(c) and Fig. 7(d), it can be inferred that with successive epochs, the gap between Train and Test Accuracy and Loss curves keeps widening, indicating that the EfficientNet B0 model is overfitting on the given dataset.

### B. VGG16 Architecture

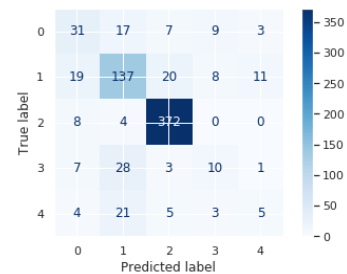

(a)

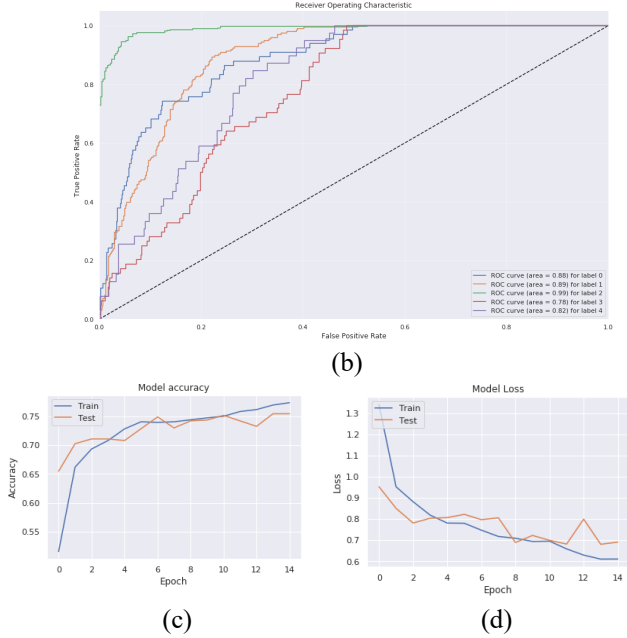

(b)

(c)           (d)

Fig. 8. (a) Confusion Matrix (b) ROC curve (c) Training and Validation Accuracy Curve (d) Training and Validation Loss Curve

From Fig. 8(a), we can observe that VGG16 perfectly classifies almost all classes except it misclassifies several Proliferate DR images as Moderate DR images, and Severe DR images also as Moderate DR images. Fig. 8(b) demonstrates that the best classification performance is achieved for Non-DR images, as it has the maximum AUC value. From Fig. 8(c) and Fig. 8(d), it can be inferred that with successive epochs, the gap between Train and Test Accuracy and Loss curves keeps depleting, indicating that the VGG16 CNN model is neither overfitting nor underfitting, and gives perfect training and validation performance on the given dataset.

*C. ResNet50 V2 Architecture*

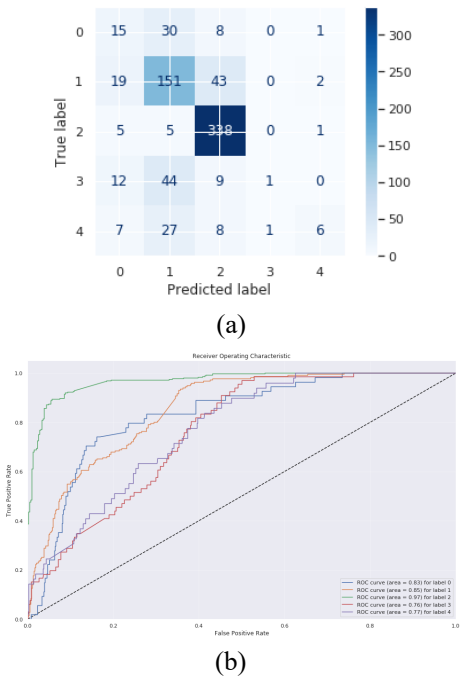

(a)

(b)

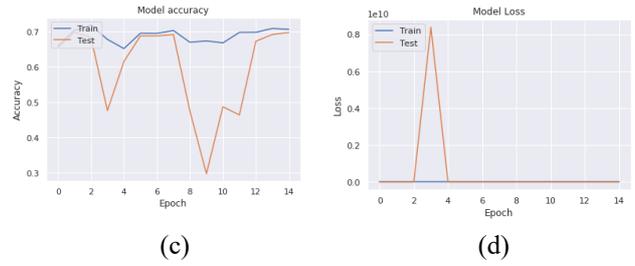

(c)           (d)

Fig. 9. (a) Confusion Matrix (b) ROC curve (c) Training and Validation Accuracy Curve (d) Training and Validation Loss Curve

From Fig. 9(a), we can observe that ResNet50 V2 misclassifies several Mild DR images as Moderate DR images, Proliferate DR images as Moderate DR images, and Severe DR images also as Moderate DR images. Fig. 9(b) demonstrates that the best classification performance is achieved for Non-DR images, as it has the maximum AUC value. From Fig. 9(c) and Fig. 9(d), it can be inferred that with successive epochs, the Train and Test Accuracy and Loss curves keep fluctuating, indicating that the ResNet50V2 model is not learning properly on the given dataset.

TABLE 1. Performance Summary

| Model Name | Performance Metrics | Mild DR | Moderate DR | No DR | Proliferate DR | Severe DR |
|---|---|---|---|---|---|---|
| VGG16 | TPR | 0.31 | 0.86 | 0.95 | 0.12 | 0.07 |
| | TNR | 0.97 | 0.77 | 0.94 | 0.98 | 0.98 |
| | PPV | 0.51 | 0.58 | 0.94 | 0.38 | 0.27 |
| | NPV | 0.93 | 0.93 | 0.95 | 0.92 | 0.95 |
| | FPR | 0.02 | 0.22 | 0.05 | 0.01 | 0.01 |
| | FNR | 0.68 | 0.13 | 0.04 | 0.87 | 0.92 |
| | FDR | 0.48 | 0.41 | 0.05 | 0.61 | 0.72 |
| | ACC | 0.91 | 0.80 | 0.95 | 0.90 | 0.93 |
| EfficientNetB0 | TPR | 0.40 | 0.67 | 0.97 | 0.35 | 0.27 |
| | TNR | 0.98 | 0.87 | 0.95 | 0.93 | 0.93 |
| | PPV | 0.7 | 0.69 | 0.95 | 0.34 | 0.16 |
| | NPV | 0.94 | 0.87 | 0.97 | 0.93 | 0.96 |
| | FPR | 0.01 | 0.12 | 0.04 | 0.06 | 0.06 |
| | FNR | 0.59 | 0.32 | 0.02 | 0.64 | 0.72 |
| | FDR | 0.3 | 0.31 | 0.04 | 0.65 | 0.83 |
| | ACC | 0.92 | 0.81 | 0.96 | 0.88 | 0.90 |
| ResNet50V2 | TPR | 0.27 | 0.70 | 0.96 | 0.01 | 0.12 |
| | TNR | 0.93 | 0.79 | 0.82 | 0.99 | 0.99 |
| | PPV | 0.25 | 0.58 | 0.83 | 0.5 | 0.6 |
| | NPV | 0.94 | 0.86 | 0.96 | 0.91 | 0.94 |
| | FPR | 0.06 | 0.2 | 0.17 | 0.001 | 0.005 |
| | FNR | 0.72 | 0.29 | 0.03 | 0.98 | 0.87 |
| | FDR | 0.74 | 0.41 | 0.16 | 0.5 | 0.4 |
| | ACC | 0.88 | 0.76 | 0.89 | 0.90 | 0.93 |

From TABLE 1, it can be observed that VGG16 achieves a maximum classification accuracy of 95% for No DR Images, the minimum classification accuracy of 80% for Moderate DR Images, with the highest false-positive rate of 22%. Whilst EfficientNet B0 achieves a maximum classification accuracy of 96% for No DR Images, its overall performance is the weakest as compared to VGG16 and ResNet50 V2. ResNet50 V2 performs better than EfficientNet B0 with the highest true positive rate of 96% for No DR Images.

## V. CONCLUSION

This paper studies the performance of several modern CNN architectures on the task of Diabetic Retinopathy Classification with the help of Transfer Learning. The classification performance is visualized by plotting Training and Validation Accuracy and Loss Curve, ROC curve, and Confusion Matrix for each CNN architecture. The research focuses on VGG16, Resnet50 V2, and EfficientNet B0 models. The analysis of architecture performance demonstrates that the best classification is achieved with Transfer Learning on VGG16 Model with the highest accuracy of 95%. It is closely followed by ResNet50 V2 architecture with an accuracy of 93%. This research shows that predictive analysis of DR from retinal images is achieved with Transfer Learning on several highly efficient and scalable Convolutional Neural Networks. Thus, the objective of studying the effect of Transfer Learning on Diabetic Retinopathy Classification is accomplished.

## VI. ACKNOWLEDGMENT

We would like to thank our Professor Ms. Rakhi Kalantri, Department of Computer Engineering, Fr. C. Rodrigues Institute of Technology, Vashi for her support and guidance.